# Improvement of Automatic Hemorrhages Detection Methods Using Shapes Recognition

Nidhal Khdhair El Abbadi[1], Enas Hamood Al Saadi[2]

[1]Computer Science Dep., University of Kufa, Najaf, Iraq
[2]Computer Science Dep., University of Babylon, Babylon, Iraq
nidhal.elabbadi@uokufa.edu.iq

## ABSTRACT

Diabetic Retinopathy is a medical condition where the retina is damaged because fluid leaks from blood vessels into the retina. The presence of hemorrhages in the retina is the earliest symptom of diabetic retinopathy. The number and shape of hemorrhages is used to indicate the severity of the disease. Early automated hemorrhage detection can help reduce the incidence of blindness.

This paper introduced new method depending on the hemorrhage shape to detect the dot hemorrhage (DH), its number, and size at early stage, this can be achieved by reducing the retinal image details. Detection and recognize the DH by following three sequential steps, removing the fovea, removing the vasculature and recognize DH by determining the circularity for all the objects in the image, finally determine the shape factor which is related to DH recognition, this stage strengthens the recognition process. The proposed method recognizes and separates all the DH.

# 1. INTRODUCTION

Nowadays medical imaging provides major aid in many branches of medicine; it enables and facilitates the capture, transmission and analysis of medical images as well as providing assistance toward medical diagnosis. Medical imaging is still on the rise with new imaging modalities being added and continuous improvements of devices capabilities (R. Rodriguez et al, 2008).

Automatic analysis of funds images of eye is an important task whose objective is to assist ophthalmologists in the diagnosis of diseases such as diabetic retinopathy and age related macular degeneration which is the main causes of blindness in several patients with these diseases. It is very important for ophthalmologists to determine these diseases in early stages because some of these diseases, if not detected early, can make people blind (A. Bessaid et al., 2009).

Physicians and ophthalmologists assess retinal images for several kinds of lesions, including hemorrhages, exudates, and arteriolar narrowing. Hemorrhage is a major sign of diabetic retinopathy, which is the second most common cause of vision loss (Y Hatanaka et al., 2011).

Diabetes, though generally non-fatal, can affect other vital organs of the human body. If not treated in the early stages, the organs affected by diabetes can malfunction or completely stop functioning.

Diabetic retinopathy (DR), as the name suggests, is a disease of the human retina caused by diabetes. It is diagnosed by observing the extent of two different kinds of defects on the retina: (1) *Micro-aneurysms* and (2) regions of *capillary non-perfusion*.

Micro-aneurysms are amongst the first signs of the presence of diabetic retinopathy. However, it is important to note that, while a critical component of any DR screening system, detection of micro-aneurysms is not equivalent to detection of DR (Meindert N et al., 2010). Micro-aneurysms are 'sprouts' of newly developing blood vessels in the retina. On the other hand, regions of capillary non-perfusion (CNP) are regions where the capillary network in the retina stops supplying blood to the corresponding areas. Micro-aneurysms are defects occurring in the early stages of DR, while CNP is a defect occurring in the later stages of DR.

As the amount of blood supplying the retina is decreased, the sight may be degraded and in the severe cases, blindness may occur. The need to develop an automated laser system to treat the whole retina in one session has become a necessity.

The detection of hemorrhages is one of the important factors in the early diagnosis of diabetic retinopathy (DR). The existence of hemorrhages is generally used to diagnose DR or hypertensive retinopathy by using the classification scheme of Scheie. In spite of detecting micro-aneurysms, it is difficult for ophthalmologists to find them in non-contrast fundus images. The contrast observed in a micro-aneurysm image is very low; therefore, ophthalmologists usually detect micro-aneurysms by using fluorescein angiograms. However, it is difficult to use fluorescein as a contrast medium for diagnosing all the medical examiners subjected to mass screening. Therefore, the patients who show the possibility of having DR were thoroughly examined at a hospital.

As new blood vessels form at the back of the eye as a part of *proliferative diabetic retinopathy* (PDR), they can bleed (vitreous hemorrhage) and blur vision. The first time this happens, it may not be very severe. In most cases, it will leave just a few specks of blood, or dots, floating in a person's visual field, though the dots often go away after a few hours. These dots are often followed within a few days or weeks by a much greater leakage of blood, which blurs vision. In extreme cases, a person will only be able to tell light from dark in that eye. It may take the blood anywhere from a few days to months or even years to clear from the inside of the eye, and in some cases the blood will not clear. These types of large hemorrhages tend to happen more than once, often during sleep.

Most of the existing methods of hemorrhage detection can be divided into two consequent stages: red lesion candidate extraction and classification. The first stage requires image preprocessing to reduce noise and improve contrast. After that the red area of the picture are extracted and segmented to be the candidate of red lesion. The vessel segmentation algorithms are applied to extract the blood vessel from the candidates to reduce false detection. Then the feature analysis which involves feature extraction and feature selection is used to detect hemorrhage.

# 2. RELATED WORKS

In 2010 Neera and Chandra presented a method to automated early detection of diabetic retinopathy using image analysis techniques. The automated diabetic retinopathy diagnosis system is thus used to various lesions of the retina i.e. exudates, micro-aneurysms and hemorrhages and their count size and location to assess the severity of the disease so that the patient can be diagnosed early and referred to the specialist well in advance for further intervention (S. Neera, and R. Chandra, 2010).

In 2011 Zhang et al. proposed a background estimation algorithm to detect hemorrhage. They saw the area of interest e.g. hemorrhages and vessels as foreground that will be visibly in contrast with background. To extract non-hemorrhage features, they introduced non-vessel inhibition operator using Gabor filter to detect vessel feature and applied a multi-thresholds scheme based on standard hysteresis thresholding methods to help separate connective elongated vessels from scattered residual edge. The background estimation used Mahalanobis distance along with a threshold technique to separate foreground from background (Zhang D et al., 2011).

Tang et al. in 2011 proposed a large hemorrhage detection method based on splat feature classification. The fundus image was segmented in to several splats of the same color based on the assumption that the pixels of the same structure have similar color and are located spatially. Each splat can be extracted as a distinct feature e.g., hemorrhages and blood vessels. A classifier was trained to recognize the splats with vessels and then used to extract the vessels from the image, leaving what considered hemorrhage candidates behind (Tang L et al., 2011).

Köse et al. in 2012 used inverse segmentation method to separate the healthy region from the unhealthy region. Based on the fact that the texture of healthy region does not vary as much as the texture of the unhealthy region, it is more accurate to extract the healthy region. Then dark lesions are left after segmentation using the intensity value that is lower than the background intensity value, and extraction of vessels (Köse C et al., 2012).

In 2013 Li Tang *et al.*, presented a method which partitions retinal color images into non overlapping segments covering the entire image. Each segment, i.e., splat, contains pixels with similar color and spatial location. A set of features is extracted from each splat to describe its characteristics relative to its surroundings, employing responses from a variety of filter bank, interactions with neighboring splats, and shape and texture information. An optimal subset of splat features is selected by a filter approach followed by a wrapper approach (Li Tang et al, 2013).

## 3. METHODOLOGY

The intensity plays an important role in the detection of dot hemorrhage, practically the "dark" part in retina image represented with low numbers in terms of intensities, each pixel in image has intensity value ranging from 0 (darkest pixel), and 255 (lighter pixel). The regions with high and low intensities in image may have very important features because it is marked as image objects. The first step of processing retain images is to reduce image details by converting color image to a gray scale image.

In an image of several objects, points of low intensity could represent the interested objects; this minimum intensity can be used to identify objects in an image. An image can have multiple low intensity, but only regions which has intensities smaller than or equal to threshold (12) (the band of red, dark, and obscured region) and has similar texture determined. This process converts the image to binary image by changing each pixel equal or smaller than threshold to white color with value 1, while the other pixel will change to black color with value 0. The threshold is determined by experiments.

At this case all the image details are changed to background as black color, while the white pixels represent the dot hemorrhage, fovea, and parts of vasculature (parts of vasculature will be removed at this stage but not all). Figure 1 shows the image in different stages.

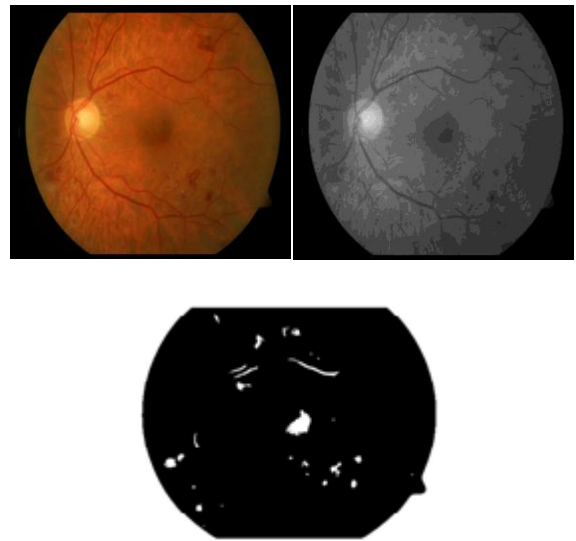

Figure 1: origin image (colored image), gray scale image, and binary image

The next step is to label all the components (regions) in the binary image by using image label algorithm as in figure 2, which scans all image pixels, assigning preliminary labels to nonzero pixels and recording label equivalences in a union-find table. Then, resolve the equivalence classes using the union-find algorithm (The Union- Find algorithm is used for maintaining a number of non-overlapping sets from a finite universe of elements). Finally, re-label the pixels based on the resolved equivalence classes.

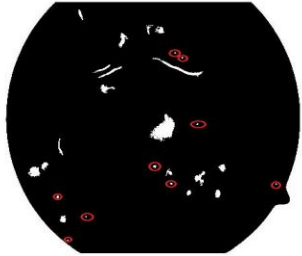

Figure 2: binary image with labeling dot hemorrhage

The final step is to recognize the dot hemorrhage from the other similar objects, by using three sequence processes:

**1.** Determine the area for each labeled region. At the early case of diabetic, the dot hemorrhage will be very small. It is very useful to recognize the dot hemorrhage at the beginning of composition.
The fovea size is about 0.46% of image size (for image with 466×489 pixels, the fovea size will be about 1067 pixels), for that almost the dot hemorrhage size will be smaller than the fovea size ( except may be for later case of diabetic, at this case every objects can be recognized by the naked eye). At this step we will segment the fovea from the other objects (dot hemorrhage, and parts of vasculature).

**2.** The second stage focuses on recognizing the dot hemorrhage from the parts of vasculatures depending on the shape of these components, the dot hemorrhage has almost a circular shape, while the vasculature does not have circular shape (the methods used to convert image to binary image, this method will leave the vasculature as it is without breaking it to small parts of vessels). Then any not circular shapes will remove like the blood vessels.
This is achieved by determine the circularity of image regions from the relation:

**Circularity = 4 × pi × (area/perimeter^2)**

Area and perimeter is the area and perimeter of the interested labeled region which counts the circularity for it.

Now, if the **circularity ≥ 1** the region is dot hemorrhage. Otherwise it isn't dot hemorrhage.

**3.** The last step used to increase the accuracy in step 2, which depends on the phenomena that the dot hemorrhage has almost a circle shape, and any not circular shape represents other thing does not represent dot hemorrhage like blood vessel.
The circle shape can be determined by using the following relation:

$$\text{shape factor (SF)} = \frac{\text{area}}{(\text{diameter})^2}$$

Area is the labeled region area, while the diameter is the maximum distance between two pixels in a labeled region. A shape factor of 0.785 is used to segment the circular features from the unwanted red regions (dot hemorrhage), (each shape factor value recognize a specific shape like SF = 0.5 used to recognize square shape, and SF= 0.4 used for rectangle...and so on).

By using these three steps, all the dot hemorrhage will be recognized and can determine its size. Figure 3 shows the dot hemorrhage detection algorithm.

## 4. CONCLUSION

In this paper we have tried to overcome the problems which accompany the converting of input image to binary image in previous works, which breaks the vasculatures to many small vessels almost with circular shape and have the same shape factor threshold as for dot hemorrhage.

Also the paper processed the dots hemorrhage three times with three different methods to recognize the dot hemorrhage without any doubt. This paper introduced new promise method to recognize the dot hemorrhage at early stages due to diabetic, and give very good result.

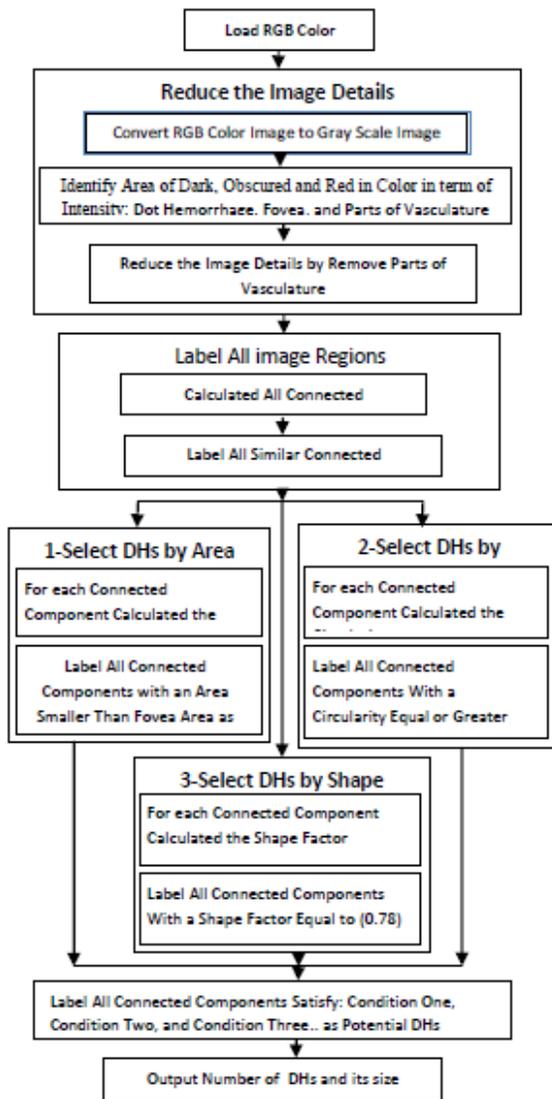

Figure 3: Dot Hemorrhage Detection Algorithm

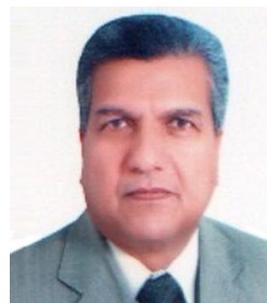

**Nidhal El Abbadi**, received BSc in Chemical Engineering, MSc, and PhD in computer science, worked in industry and many universities, he is general secretary of colleges of computing and informatics society in Iraq, Member of Editorial board of Journal of Computing and Applications, reviewer for a number of international journals, has many published papers and three published books (Programming with Pascal,

C++ from beginning to OOP, Data structures in simple language), his research interests are in image processing, biomedical, and steganography, He's Associate Professor in Computer Science in the University of Kufa – Najaf, IRAQ.

**Enas Hamood**, received her BSc. in Computer Science from Babylon University, Msc. in Computer Science from Babylon University, currently she is a PhD student in Computer Science Department, college of science- Babylon University.
She has worked as a lecturer in College of Education / Babylon University. She has done 6 researches in computer science fields.